\pdfoutput=1
\documentclass[11pt]{article}

\usepackage{acl}

\usepackage{times}
\usepackage{latexsym}

\usepackage[T1]{fontenc}
\usepackage[utf8]{inputenc}

\usepackage{microtype}
\usepackage{bm}
\usepackage{mathtools}
\usepackage{multirow}
\usepackage{graphicx}
\usepackage{xhfill}
\usepackage{amssymb}
\usepackage{here}
\usepackage{pifont}
\usepackage[subrefformat=parens]{subcaption}
\usepackage {arydshln}
\usepackage{cleveref}

\newcommand{\figlabel}[1]{\label{fig:#1}}
\newcommand{\figref}[1]{Figure \ref{fig:#1}}
\newcommand{\tbllabel}[1]{\label{tbl:#1}}
\newcommand{\tblref}[1]{Table \ref{tbl:#1}}
\newcommand{\seclabel}[1]{\label{sec:#1}}
\newcommand{\secref}[1]{Section \ref{sec:#1}}
\newcommand{\equlabel}[1]{\label{eq:#1}}
\newcommand{\equref}[1]{Eq. (\ref{eq:#1})}

\newcommand{\mr}[2]{\multirow{#1}{*}{#2}}
\newcommand{\ml}[3]{\multicolumn{#1}{#2}{#3}}

\newcommand{\bhline}[1]{\noalign{\hrule height #1}}

\title{Post-processing Networks: Method for Optimizing Pipeline Task-oriented Dialogue Systems using Reinforcement Learning}

\author{
Atsumoto Ohashi\qquad Ryuichiro Higashinaka \\
Graduate School of Informatics, Nagoya University\\
\texttt{ohashi.atsumoto.c0@s.mail.nagoya-u.ac.jp} \\
\texttt{higashinaka@i.nagoya-u.ac.jp} \\
}

\begin{document}
\maketitle
\begin{abstract}
Many studies have proposed methods for optimizing the dialogue performance of an entire pipeline task-oriented dialogue system by jointly training modules in the system using reinforcement learning. However, these methods are limited in that they can only be applied to modules implemented using trainable neural-based methods. To solve this problem, we propose a method for optimizing a pipeline system composed of modules implemented with arbitrary methods for dialogue performance. With our method, neural-based components called post-processing networks (PPNs) are installed inside such a system to post-process the output of each module. All PPNs are updated to improve the overall dialogue performance of the system by using reinforcement learning, not necessitating each module to be differentiable. Through dialogue simulation and human evaluation on the MultiWOZ dataset, we show that our method can improve the dialogue performance of pipeline systems consisting of various modules\footnote{Our code is publicly available at \url{https://github.com/nu-dialogue/post-processing-networks}}.
\end{abstract}

\section{Introduction}
Task-oriented dialogue systems can be classified into two categories: pipeline systems, in which multiple modules take on a sequential structure, and neural-based end-to-end systems \citep{chen2017survey, gao2018neural, zhang2020recent}.

A typical pipeline system consists of four modules \citep{zhang2020recent}: natural language understanding (NLU), dialogue state tracking (DST), Policy, and natural language generation (NLG). Each module can be implemented individually using various methods (e.g., rule-based and neural-based) \citep{ultes-etal-2017-pydial, zhu-etal-2020-convlab}. In a pipeline system, the inputs and outputs of each module are explicit, making it easy for humans to interpret. However, since each module is processed sequentially, errors in the preceding module can easily propagate to the following ones, and the performance of the entire system cannot be optimized \citep{tseng-etal-2021-transferable}. This results in low dialogue performance of the entire system \citep{takanobu-etal-2020-goal}.

In contrast, neural-based methods can optimize entire neural-based end-to-end systems, which allows for less error propagation than pipeline systems and high dialogue performance \citep{dinan2019second, gunasekara2020overview}. The drawback of these methods is the large amount of annotation data required to train systems \citep{zhao-eskenazi-2016-towards}. Compared with pipeline systems, neural-based end-to-end systems are also less interpretable and more difficult to adjust or add functions.

\begin{figure}[t]
\begin{minipage}[t]{\linewidth}
\centering
\includegraphics[scale=0.173]{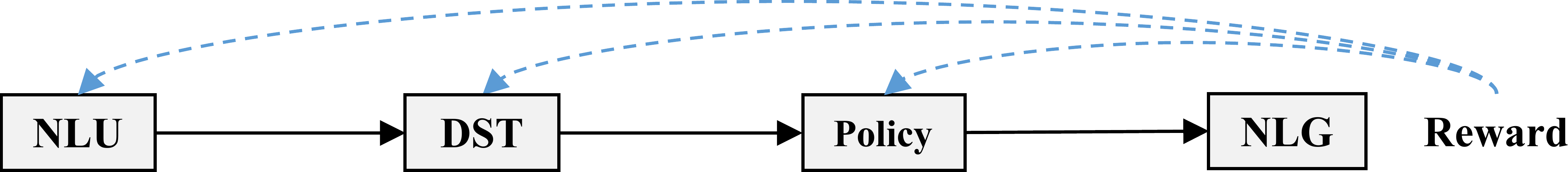}
\subcaption{Diagram of conventional method. Modules are fine-tuned using RL.}
\figlabel{conventional_method}
\end{minipage} \\
\begin{minipage}[t]{\linewidth}
\centering
\includegraphics[scale=0.173]{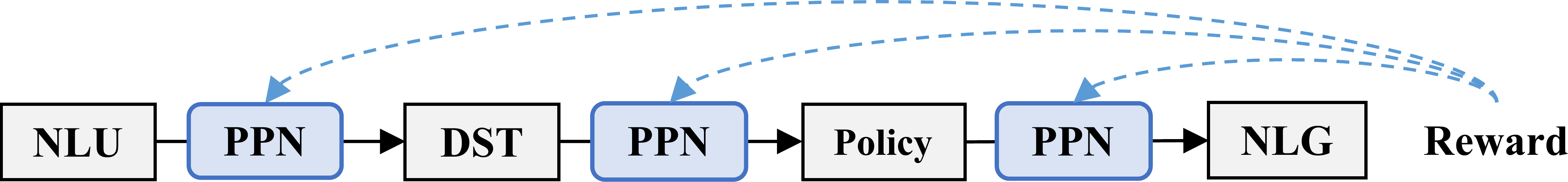}
\subcaption{Diagram of proposed method. Each PPN that post-processes output of each module is optimized using RL.}
\figlabel{proposed_method}
\end{minipage} \\
\caption{Comparison of conventional and proposed methods}
\figlabel{rl_methods_comparison}
\vspace{-3mm}
\end{figure}

To marry the benefits of both pipeline and end-to-end systems, methods \citep{liu-etal-2018-dialogue, mehri-etal-2019-structured, 9514885, lin2021joint} have been proposed for optimizing an entire pipeline system in an end-to-end fashion by using reinforcement learning (RL) (\figref{conventional_method}). These methods are powerful because they jointly train and fine-tune neural-based implementations of the modules, such as NLU, Policy, and NLG, by using RL. However, these methods may not always be applicable because there may be situations in which modules can only be implemented with rules or the modules’ internals cannot be accessed, such as with a Web API.

With this background, we propose a method for optimizing an entire pipeline system composed of modules implemented in arbitrary methods. We specifically focus on modules that output fixed sets of classes (i.e., NLU, DST, and Policy) and install neural-based components (post-processing networks; PPNs) in the system to post-process the outputs of these modules, as shown in \figref{proposed_method}. Each PPN modifies the output of each module by adding or removing information as necessary to facilitate connections to subsequent modules, resulting in a better flow of the entire pipeline. To enable the appropriate post-processing for the entire system, each PPN uses the states of all modules in the system when executing post-processing. The post-processing of each PPN is optimized using RL so that the system can improve its dialogue performance, e.g., task success. A major advantage of our method is that each module does not need to be trainable since PPNs are trained instead.

To evaluate the effectiveness of our method, we applied PPNs to pipeline systems consisting of modules implemented with various methods (e.g., rule-based and neural-based) on the basis of the MultiWOZ dataset \citep{budzianowski-etal-2018-multiwoz} and conducted experiments by using dialogue simulation and human participants. The contributions of this study are as follows.
\begin{itemize}
\item We propose a method of improving the dialogue performance of a pipeline task-oriented dialogue system by post-processing outputs of modules. Focusing on NLU, DST, and Policy, our method can be applied to various pipeline systems because PPNs do not depend on the implementation method of each module or a combination of modules.
\item Dialogue simulation experiments have shown that our method can improve the dialogue performance of pipeline systems consisting of various combinations of modules. Additional analysis and human evaluation experiments also verified the effectiveness of the proposed method.
\end{itemize}

\section{Related Work}
Our study is related to optimizing an entire dialogue system with a modular architecture. \citet{wen-etal-2017-network} proposed a method for implementing all the functions of NLU, DST, Policy, and NLG modules by using neural networks, enabling the entire system to be trained. \citet{lei-etal-2018-sequicity} incorporated both a decoder for generating belief states (i.e., DST module) and a response-generation decoder (i.e., NLG module) into a sequence-to-sequence model \citep{sutskever2014sequence}. \citet{Zhang_Ou_Yu_2020} also proposed a method for jointly optimizing a system that includes three decoders that respectively execute the functions of DST, Policy, and NLG. \citet{Liang_Tian_Chen_Yu_2020} extended the method of \citet{lei-etal-2018-sequicity} by jointly optimizing four decoders that generate user dialogue acts (DAs), belief states, system DAs, and system responses. However, these systems are trained in a supervised manner and require large amounts of data \citep{liu2017end}.

Our study is related to improving the dialogue performance of a pipeline system by using RL. \citet{zhao-eskenazi-2016-towards} and \citet{li-etal-2017-end} implemented DST and Policy in a neural model and used the Deep Q Network \citep{mnih2013playing} algorithm to optimize the system to achieve robustness against errors that occur in interactions. \citet{liu-etal-2018-dialogue} proposed a Policy-learning optimization method for real users by combining supervised learning, imitation learning, and RL. \citet{mehri-etal-2019-structured} proposed a method for training a response-generation model by using RL while using the hidden states of the learned NLU, Policy, and NLG. Methods have been proposed \citep{9514885, lin2021joint} for building a pipeline system with individually trained modules and fine-tuning specific modules by using RL, which significantly improved the performance of the overall system. These methods are powerful because they can fine-tune a system directly through RL. However, they can only be applied to systems consisting of specific differentiable modules implemented using neural-based methods, not to systems consisting of non-differentiable modules. Our method is independent of the module-implementation method, trainability of each module in pipeline systems, and combination of modules.

\begin{figure*}[t]
\centering
\includegraphics[scale=0.132]{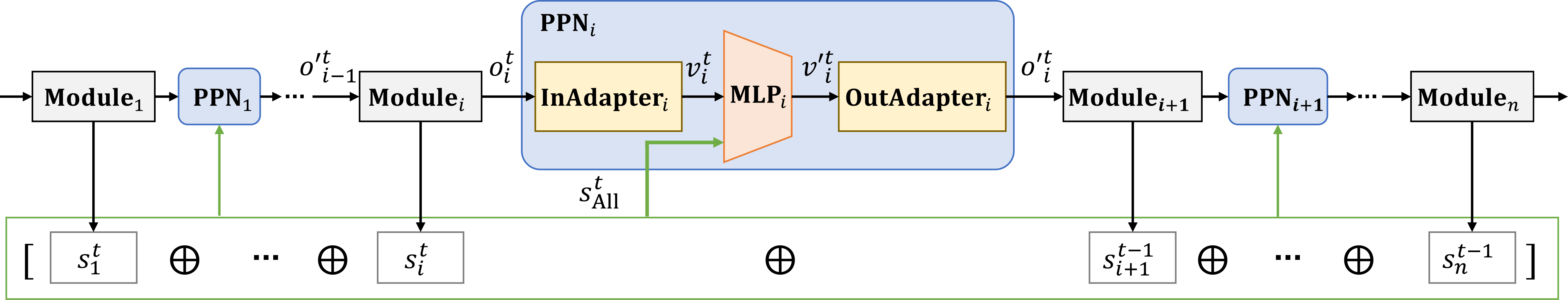}
\caption{Architecture of our proposed method. Output of each module is post-processed by subsequent PPN. Each PPN has InAdapter to convert output label $o$ of module into multi-binary vector $v$, MLP to post-process multi-binary vector into $v'$ on basis of $v$ and state $s_{\rm{All}}$ of all modules, and OutAdapter to restore $v'$ to output label $o'$.}
\figlabel{ppn_architecture}
\vspace{-3mm}
\end{figure*}

\begin{figure}[t]
\centering
\includegraphics[scale=0.1]{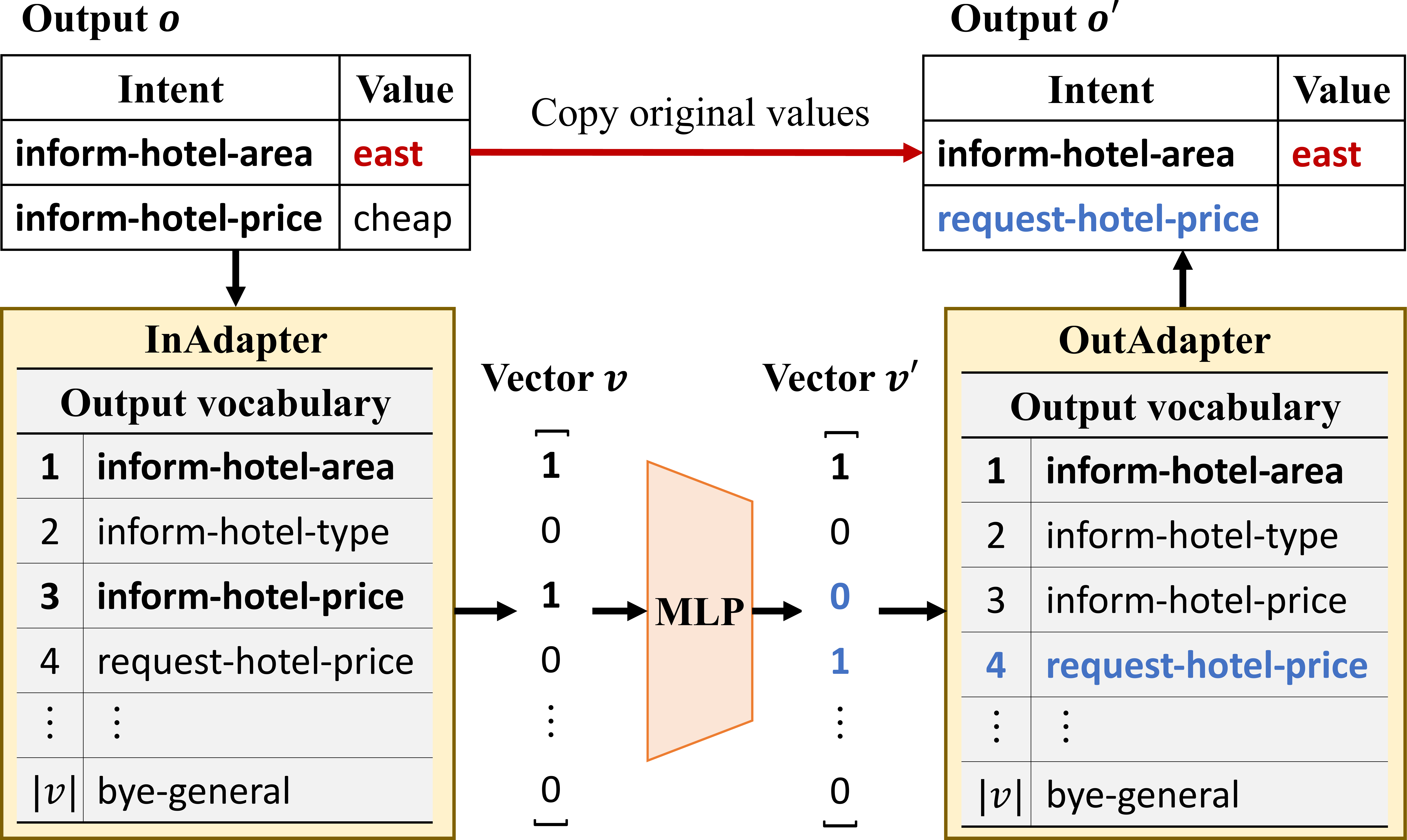}
\caption{Procedure in which InAdapter converts output label $o$ into vector $v$ and OutAdapter restores vector $v'$ into output label $o'$ by using output vocabulary (in this case, output labels are DAs of NLU). Value information, which cannot be encoded in $v$, is copied directly from $o$ when creating $o'$.}
\figlabel{adapter_example}
\vspace{-3mm}
\end{figure}

\section{Proposed Method}
\seclabel{method}
We developed our method to improve the dialogue performance of an entire pipeline system by optimizing the output of each module through post-processing. Post-processing means modifying the output by adding or removing information from the actual output of the module. With our method, each PPN needs to execute post-processing appropriate for all modules so that the entire system can improve overall dialogue performance. With this in mind, each PPN post-processes the target module's output while using the latest states of all modules in the system. Basically, each module's state is the latest output of each module. However, if a module can provide information that represents its state in more detail than the module's output, the PPN also uses that information (see \secref{post_processing_algorithm}). \figref{ppn_architecture} shows the architecture of PPNs applied to a pipeline system consisting of Module$_1$, ..., Module$_n$. 

\subsection{Post-processing Algorithm}
\seclabel{post_processing_algorithm}
The following equations describe the steps in which PPN$_i$ post-processes the output $o^t_i$ of Module$_i$ at turn $t$, as in \figref{ppn_architecture}.

\begin{eqnarray}
o^t_i, s^t_i &=& {\rm Module}_i(o'^t_{i-1}) \equlabel{1}\\
v^t_i &=& {\rm InAdapter}_i(o^t_i) \equlabel{2}\\
s^t_{\rm{All}} &=& [s^t_1;...;s^t_i;s^{t-1}_{i+1};...;s^{t-1}_n] \equlabel{3}\\
v'^t_i &=& {\rm MLP}_i([v^t_i; s^t_{\rm{All}}]) \equlabel{4}\\
o'^t_i &=& {\rm OutAdapter}_i(v'^t_i) \equlabel{5}
\end{eqnarray}

As in a general pipeline system, Module$_i$ first receives the output $o'^t_{i-1}$ of the preceding Module$_{i-1}$ and outputs $o^t_i$ as the result of its processing (\equref{1}) (e.g., for the NLU module, it receives the user's utterance as input and outputs the user's DAs). At the same time, Module$_i$ outputs its additional information $s^t_i$ obtained in the processing, which is related to the state of Module$_i$ (\equref{1}). Basically, $s^t_i$ is the same as $o^t_i$. However, if Module$_i$ can provide more detailed information about its state obtained in the processing (e.g., for the NLU module, it typically outputs confidence scores of predicted user's DAs), Module$_i$ outputs that information as $s^t_i$.

Next, $o^t_i$ is input to PPN$_i$. In PPN$_i$, InAdapter$_i$ creates a multi-binary vector $v^t_i$, which is a vector representation of $o^t_i$ (\equref{2}). The left half of \figref{adapter_example} shows a concrete example of an InAdapter converting a module output into a multi-binary vector. The InAdapter$_i$ is created by hand-crafted rules using the output vocabulary set of Module$_i$. At the same time as creating $v^t_i$, $s^t_{\rm{All}} = [s^t_1; ... ;s^t_i; s^{t-1}_{i+1}; ... ;s^{t-1}_n]$, which is a concatenation of the latest states of Module$_1$, ..., Module$_n$, are also created (\equref{3}). Note that $s^{t-1}$ is used for states of Module$_{i+1}$, ..., Module$_n$ because modules after Module$_i$ have not produced their states in turn $t$.

The $v^t_i$ and $s^t_{\rm{All}}$ created thus far are combined and input to multi-layer perceptron (MLP) MLP$_i$, which outputs a multi-binary vector $v'^t_i$ (\equref{4}). The dimensions of $v'^t_i$ are the same as the vocabulary size of Module$_i$. At this point, the changes in the original vectors $v^t$ and $v'^t$ become the result of post-processing. That is, the dimension, the value in $v^t_i$ of which is $1$ and value in $v'^t_i$ of which is $0$, is the information deleted by MLP$_i$, and the reverse is the information added by MLP$_i$. Finally, OutAdapter$_i$ converts $v'^t_i$ into $o'^t_i$, the output label representation of Module$_i$. Some of the value information is directly copied from $o^t_i$ when creating $o'^t_i$ since these values are not given by $v'^t_i$. If there is no need to fill in the value, it is left empty. The right half of \figref{adapter_example} shows a concrete example of an OutAdapter converting a multi-binary vector into a label representation of a module's vocabulary. As with InAdapter$_i$, OutAdapter$_i$ is created by hand-crafted rules using the output vocabulary set of Module$_i$. 

At runtime, in the initial turn, the states of some modules that have never processed yet are initialized with zero vector (i.e., $s^0=\bm{0}$). In the subsequent turn $t$, as mentioned above, PPN$_i$ uses the preceding modules' states $[s^t_1, ..., s^t_i]$ and the succeeding modules' states $[s^{t-1}_{i+1}, ..., s^{t-1}_n]$.

With our method, the MLPs of all PPNs are optimized jointly by using RL via interaction with users (see \secref{rl_training}). To apply PPNs to a system, we only need the vocabulary set of each module to implement an InAdapter and OutAdapter for conversion. Therefore, our method can be applied to both differentiable and non-differentiable implementations of the modules. Since we want first to verify the idea of PPNs, we only used MLPs and focused on NLU, DST, and Policy in this study. Once the verification is complete, we aim to apply PPNs to more complex modules, such as NLG.

\subsection{Pre-training with Imitation Learning}
\seclabel{pre_training}
It is not easy to optimize an MLP from scratch by using RL. Many studies have shown that model performance can be improved by imitation learning, which is a scheme for learning to imitate the behavior of experts before RL is conducted \citep{12e2908dea004005a257707a1ad43237, rajeswaran2017learning}. We considered the actual output $o_i$ of Module$_i$ to be the behavior of the expert for PPN$_i$ and conducted supervised learning so that PPN$_i$ copies $o^t$ before RL. This should allow each PPN to focus only on ``how to modify the module's output $o$'' during RL.

With our method, a pipeline system consisting of Module$_1$,..., Module$_n$ first executes dialogue sessions for sampling training data. In each dialogue, we sample the $[s_{\rm{All}}, v]$ of each module for all turns. At this stage, no PPNs execute post-processing, and no MLPs are used. When training MLPs by imitation learning, supervised learning is carried out using the sampled data. We train all MLPs to execute a multi-labeling task in which the input is $[v; s_{\rm{All}}]$ and the output label is $v$. Binary cross-entropy is used to update the MLP to minimize the difference between $v$ and $v' = {\rm MLP}([v; s_{\rm{All}}])$.

\subsection{Optimization with Reinforcement Learning}
\seclabel{rl_training}
The goal with PPNs is to improve dialogue performance (e.g., task success) by each PPN post-processing the output of each module. Therefore, the MLP of each PPN needs to be optimized using RL for maximizing the rewards related to dialogue performance. We use proximal policy optimization (PPO) \citep{schulman2017proximal} as the RL algorithm, which is a stable and straightforward policy-gradient-based RL algorithm.

The following steps show the learning algorithm of a PPN for each iteration:
\begin{description}
\item[Step. 1] The pipeline system with PPNs interacts with a user. Each PPN post-processes and samples the $s^t_{\rm{All}}$, $v^t$, $v'^t$, and reward $r^t$ of each MLP in turn $t$. The sampled $\lparen s^t_{\rm{All}}, v^t, v'^t, r^t \rparen$ are added to the post-processing history (called \emph{trajectory}) of each PPN. As an $r^t$, we give the same value to all PPNs. These trials are repeated until the trajectory reaches a predetermined size (called \emph{horizon}).
\item[Step. 2] The PPN to be updated in this iteration is selected on the basis of the \emph{PPN-selection strategy}, which is a rule for selecting PPNs to be updated in each iteration. We have three strategies described in the next paragraph.
\item[Step. 3] The MLPs of the PPNs selected in Step. 2 are updated using the PPO algorithm. Each MLP is updated for multiple epochs using the trajectory sampled in Step. 1 as training data.
\end{description}

Since it is not apparent which modules' PPN should be updated and in what order, we prepared the following three PPN-selection strategies: ALL (select all PPNs in every iteration), RANDOM (randomly select one or more PPNs in each iteration), and ROTATION (select one PPN at each iteration in order). In the following experiments, we examined which strategy is the best.

\section{Experiments}
To confirm the effectiveness of our method, we applied PPNs to several different pipeline systems and evaluated dialogue performance using dialogue simulation. We also carried out a human evaluation.

\subsection{Dataset}
\seclabel{dataset}
We evaluated PPNs using modules and a user simulator implemented using the MultiWOZ dataset \citep{budzianowski-etal-2018-multiwoz}, which is a task-oriented dialogue dataset between a clerk and tourist at an information center. MultiWOZ contains 10,438 dialogues; one to three domains (seven domains in total in the dataset) appear simultaneously in each dialogue.

\subsection{Platform and User Simulator}
\seclabel{platform_and_user_simulator}
ConvLab-2\footnote{\url{https://github.com/thu-coai/ConvLab-2}} \citep{zhu-etal-2020-convlab} is a platform for multi-domain dialogue systems, which provides pre-implemented models of each module in the pipeline system and tools for end-to-end evaluation of the dialogue system.

We used the user simulator implemented in ConvLab-2. The simulator interacts with the dialogue system in natural language on the basis of the user goal given for each dialogue session. The simulator consists of a BERT \citep{devlin-etal-2019-bert}-based NLU \citep{chen2019bert}, an agenda-based Policy \citep{schatzmann-etal-2007-agenda}, and a template-based NLG. The agenda-based Policy models a user's behavior in MultiWOZ by using a stack-like agenda created using hand-crafted rules. A user goal for each dialogue is randomly generated: the domains are randomly selected from one to three domains (out of all seven domains) on the basis of the domains' frequency in MultiWOZ; the slots are also randomly selected on the basis of the slots' frequency in MultiWOZ.

\subsection{Evaluation Metrics}
\seclabel{evaluation_metrics}
In evaluating each dialogue, we used the number of turns\footnote{One user utterance and its system response form one turn.} (Turn) to measure the efficiency of completing each dialogue; the smaller the Turn is, the better the system performance. We also measured whether the system responds to the requested slot by the user without excess or deficiency (Inform F1) and whether the entity presented by the system met the condition of the user goal (Match Rate). We also used Task Success as a result of Match Rate and Inform Recall being equal to 1 within 20 turns. The above four metrics are the major ones for dialogue evaluation and have been used in many studies using ConvLab-2 \citep{li-etal-2020-guided, takanobu-etal-2020-goal, hou-etal-2021-imperfect}.

\subsection{Implementation}
\seclabel{implementation}
\subsubsection{System Configurations}
\seclabel{system_configuration}
To select the modules that make up a pipeline system, we referred to \citet{takanobu-etal-2020-goal}, who developed and evaluated various combinations of modules using ConvLab-2. For the models of each module (NLU, DST, Policy, and NLG), we included both classical rule-based and recent neural-based models. Note that, since this study focused on whether PPNs can be used to optimize pipeline systems consisting of non-trainable modules, we did not update modules even if the modules may be trainable. Each of the models\footnote{For models, we used the best ones provided by ConvLab-2 as of October 20, 2021} we prepared are as follows.

\paragraph{NLU} We used BERT NLU \citep{chen2019bert} for the NLU module. This model estimates DAs by tagging which domain-intent-slot each token in a user utterance represents by using a pre-trained BERT \citep{devlin-etal-2019-bert}. The InAdapter/OutAdapter are created using the DA set defined in BERT NLU (see \figref{adapter_example} for an illustration of an InAdapter-processing example by using a DA set). We used the estimated probabilities of each DA as BERT NLU's state $s$.

\paragraph{DST} We used two models for the DST module: Rule DST \citep{zhu-etal-2020-convlab} and TRADE \citep{wu-etal-2019-transferable}. Rule DST updates the dialogue state consisting of belief state, database search results, current user DAs, and previous system DAs at each turn by directly using the DAs estimated by the NLU. On the contrary, TRADE is a neural-based model that directly extracts slot-value pairs and generates belief states using the dialogue history as input. For DST modules, a belief state is subject to post-processing. Therefore, we created an InAdapter/OutAdapter on the basis of the slot types defined in the belief state on ConvLab-2. As states of Rule DST and TRADE, an entire dialogue state is converted into a multi-binary vector by using a vectorizer implemented in ConvLab-2.

\paragraph{Policy} We used four models for the Policy module: Rule Policy \citep{zhu-etal-2020-convlab}, MLE Policy, PPO Policy \citep{schulman2017proximal}, and LaRL Policy \citep{zhao-etal-2019-rethinking}. Rule Policy is a model based on hand-crafted rules. MLE Policy is a model trained on state-action pairs in MultiWOZ using supervised learning. PPO Policy is a fine-tuned model based on MLE Policy using the PPO RL algorithm. Unlike the other Policy models, LaRL Policy is an LSTM-based model trained to directly generate system utterances instead of system DAs by using RL. We created an InAdapter/OutAdapter using the DA set defined in each model. For states of MLE Policy and PPO Policy, we used the estimated probability of each DA. For Rule Policy's state, we used a binary vector representation of DAs. Since the output of LaRL is a natural language, it was not subject to post-processing in this study.

\paragraph{NLG} We used two models for the NLG module: Template NLG and SC-LSTM \citep{wen-etal-2015-semantically}. Template NLG creates system responses by inserting values into templates of utterances manually created in advance for each DA. SC-LSTM is an LSTM-based model that generates utterances on the basis of DAs. For the same reason as for LaRL Policy, we did not implement PPNs for Template NLG and SC-LSTM in these experiments.

\vskip\baselineskip
\tblref{module_dimensions} shows the dimensions of each module's state $s$ described above and the number of dimensions of the multi-binary vector $o$ of each PPN (i.e., the vocabulary of each module). Note that for the DST modules, the dimensions of $s$ and $v$ are different. This is because $s$ is a vector representation of a dialogue state, which includes a belief state, database search results, user's DAs, etc., and $v$ is a vector representation of a belief state only.

\renewcommand{\arraystretch}{1.05}
\begin{table}[t]
\small
\centering
\begin{tabular}{clcc} \bhline{0.8pt}
\textbf{Module} & \textbf{Models} & $|s|$ & $|v|$ \\ \hline
NLU & BERT & 175 & 175 \\
DST & Rule, TRADE & 340 & 24 \\
\mr{2}{Policy} & Rule, MLE, PPO & 209 & 209 \\
& LaRL & 0 & 0 \\
NLG & Template, SC-LSTM & 0 & 0 \\ \bhline{0.8pt}
\end{tabular}
\caption{Dimensions $|s|$ of state $s$ output from each module and $|v|$ of vector $v$ processed by PPN of each module. Number of output vocabularies defined for each module and $|v|$ are equal.}
\tbllabel{module_dimensions}
\end{table}

\subsubsection{Training}
Throughout all experiments, the data used for imitation learning of each pipeline system was sampled by simulating 10,000 turns, corresponding to approximately 1,000 dialogue sessions. In RL for each system, we trained 200 iterations, where one iteration consists of approximately 100 dialogue sessions. Following \citet{takanobu-etal-2019-guided}, we gave a reward of $-1$ for each turn, and when the task was a success, we gave the maximum number of turns $\times$ 2 at the end of the dialogue session, i.e., 40 in our case. See \secref{ppn_training_details} of the appendix for more training details.

To test each system, we ran 1,000 dialogues using a system that achieved the best Task Success during the RL training. Throughout all experiments, we trained with five different random seeds and reported the average of their scores as the final performance.

\begin{table}[t]
\small
\centering
{\tabcolsep=1.6mm
\begin{tabular}{lcccc} \bhline{0.8pt}
\begin{tabular}{c}{\bf PPN-selection}\\{\bf strategy}\end{tabular} & \textbf{Success} & \textbf{Inform} & \textbf{Match} & \textbf{Turn}\\ \hline
ALL & 64.2 & {\bf 71.9} & 76.6 & 9.20 \\
RANDOM & {\bf 66.1} & 71.5 & {\bf 78.7} & {\bf 8.61} \\
ROTATION & 60.4 & 70.5 & 73.2 & 9.10 \\ \bhline{0.8pt}
\end{tabular}
}
\caption{Performance after PPN training with each PPN-selection strategy}
\tbllabel{ppn_select_strategy_results}
\end{table}

\begin{figure*}[t]
\centering
\includegraphics[scale=0.35]{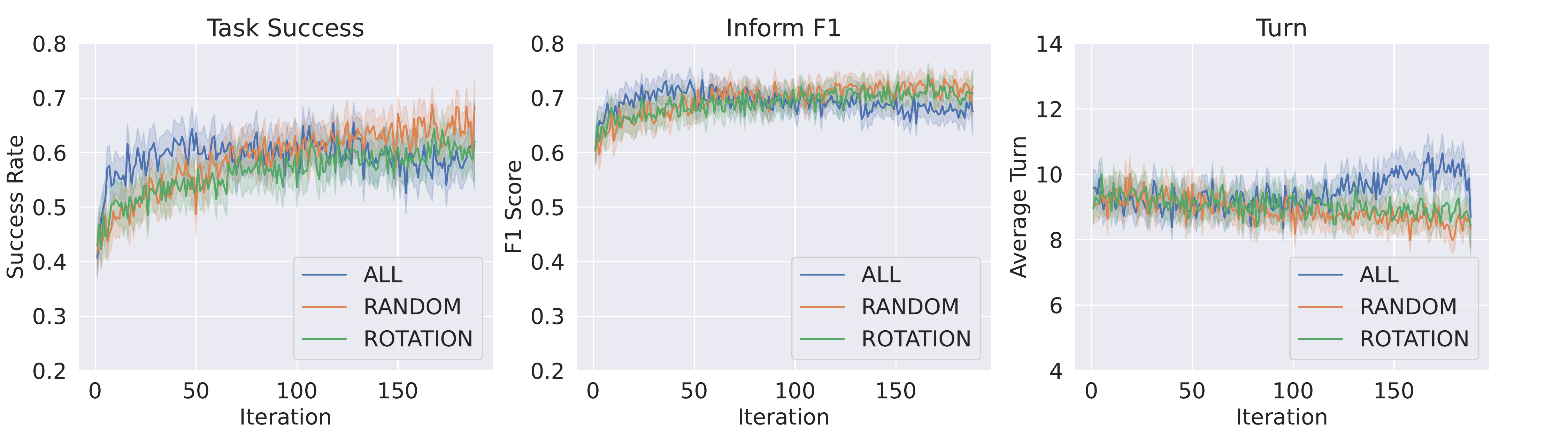}
\caption{Scores of each evaluation metric in learning process with three PPN-selection strategies}
\figlabel{train_log}
\end{figure*}

\begin{table*}[t]
\small
\centering
{\tabcolsep=2mm
\begin{tabular}{c|cccc|c|cccc} \bhline{0.8pt}
\mr{2}{\textbf{System}} & \ml{4}{|c|}{\textbf{Model Combination}} & \mr{2}{\textbf{w/ PPN}} & \mr{2}{\textbf{Task Success}} & \mr{2}{\textbf{Inform F1}} & \mr{2}{\textbf{Match Rate}} & \mr{2}{\textbf{Turn}} \\
& NLU & DST & Policy & NLG & & & & & \\ \hline
\mr{2}{SYS-RUL} & \mr{2}{BERT} & \mr{2}{Rule} & \mr{2}{Rule} & \mr{2}{Template} & & 84.1 & 87.4 & 90.2 & 5.92 \\ 
& & & & & \checkmark & 84.0 & 86.3 & {\bf 92.4} & 6.33 \\ \hline
\mr{2}{SYS-MLE} & \mr{2}{BERT} & \mr{2}{Rule} & \mr{2}{MLE} & \mr{2}{Template} & & 43.3 & 62.4 & 27.8 & 9.03 \\ 
& & & & & \checkmark & {\bf 66.1} & {\bf 71.5} & {\bf 78.7} & {\bf 8.61} \\ \hline
\mr{2}{SYS-PPO} & \mr{2}{BERT} & \mr{2}{Rule} & \mr{2}{PPO} & \mr{2}{Template} & & 54.9 & 65.5 & 55.2 & 8.41 \\ 
& & & & & \checkmark & {\bf 68.8} & {\bf 72.1} & {\bf 77.8} & {\bf 8.37} \\ \hline
\mr{2}{SYS-SCL} & \mr{2}{BERT} & \mr{2}{Rule} & \mr{2}{Rule} & \mr{2}{SC-LSTM} & & 38.3 & 57.5 & 56.7 & 13.53 \\ 
& & & & & \checkmark & {\bf 44.2} & {\bf 71.7} & {\bf 71.8} & {\bf 11.04} \\ \hline
\mr{2}{SYS-TRA} & \ml{2}{|c}{\mr{2}{TRADE}} & \mr{2}{Rule} & \mr{2}{Template} & & 19.0 & 45.6 & 36.4 & 12.08 \\ 
& & & & & \checkmark & 18.8 & {\bf 49.2} & 31.6 & 12.14 \\ \hline
\mr{2}{SYS-LAR} & \mr{2}{BERT} & \mr{2}{Rule} & \ml{2}{c|}{\mr{2}{LaRL}} & & 21.6 & 44.9 & 27.6 & 13.24 \\ 
& & & & & \checkmark & {\bf 23.9} & {\bf 50.9} & {\bf 34.1} & {\bf 12.77} \\ \bhline{0.8pt}
\end{tabular}
}
\caption{Combination of models for each pipeline system and scores before and after applying PPNs to each system. `w/ PPN' indicates whether PPNs are applied to the system. Scores that have been improved using PPNs are in bold.}
\tbllabel{model_combination_results}
\vspace{-3mm}
\end{table*}

\subsection{Experimental Procedure}
\seclabel{experimental_procedure}
We conducted four experiments. The first experiment was conducted to determine which of the PPN-selection strategies (see \secref{rl_training}) is appropriate. We used a combination of BERT NLU, Rule DST, MLE Policy, and Template NLG as the system configuration. The reasons for using this combination are that (1) the Task Success of a system composed of this module combination is around 50\%. Therefore, it would be easy to understand the impact of the PPNs, and (2) MLE Policy is used as the initial weight in many RL methods \citep{takanobu-etal-2019-guided, li-etal-2020-guided}, making it a reasonable starting point for RL. The second experiment was conducted to verify whether the PPNs work for any combination of modules; we combined some of the modules described in \secref{system_configuration} to build pipeline systems and applied PPNs. The third experiment was conducted to investigate the contribution of the PPN of each module and $s_{\rm{All}}$ to the overall performance of the system. The final experiment was a human evaluation; we examined whether the proposed method is effective not only for a simulator but also for humans.

\subsection{Comparison of Post-processing-network-selection Strategies}
\seclabel{comparison_of_ppn_selection_strategy}
\figref{train_log} shows the learning process in the three PPN-selection strategies. Task Success and Inform F1 at 50 iterations show that ALL reached the highest score about 100 iterations earlier than RANDOM and ROTATION. This is a reasonable result since the number of updates for each MLP in ALL was up to four times that for the other strategies. However, it was unstable after 50 iterations, and the scores of Task Success, Inform F1, and Turn all worsened as the learning process progressed. This is probably because the gradients of each MLP were calculated simultaneously in the PPO update algorithm, which caused each MLP to update in a different gradient direction, making it difficult for each MLP to coordinate with one another.

Although the learning speed of ROTATION and RANDOM was slow, all metrics consistently improved. Turn and Inform F1 also showed stable improvements compared with ALL. For RANDOM and ROTATION, each MLP computed its gradient after the other MLPs computed and updated their gradients one by one, which probably prevented significant discrepancies among MLPs and stabilized learning.

\tblref{ppn_select_strategy_results} shows the final performance of each strategy. RANDOM outperformed ALL in all the final scores, and ROTATION was inferior to ALL in Task Success, Inform F1, and Match Rate. Since the learning was stable and the final performance was generally better than the other strategies, we decided to use RANDOM in the following experiments.

\subsection{Comparison of Model Combinations}
\seclabel{comparison_of_module_combination}
We built six pipeline systems with different model combinations. \tblref{model_combination_results} summarizes the comparison of the scores when PPNs were applied to each system. For a fair comparison, systems without PPNs were also evaluated on the average scores\footnote{Although we used the latest models implemented in ConvLab-2, we could not reproduce the scores reported in \url{https://github.com/thu-coai/ConvLab-2\#end-to-end-performance-on-multiwoz}} of 1,000 dialogues conducted with five different random seeds.

\tblref{model_combination_results} shows that Task Success improved for most of the systems. In addition, all systems improved in Inform F1 or Match Rate. These results indicate that post-processing with PPNs can improve the dialogue performance of a pipeline system without touching the module internals. However, neither Task Success nor Turn improved for SYS-RUL and SYS-TRA. The common feature of these two systems is that they use Rule Policy and Template NLG. These modules are carefully designed by hand and originally have high accuracy, leading to little room for improvement in this configuration.

In general, there were large differences in performance among the systems regardless of whether PPN was used. As mentioned above, this is due to the performance differences among the modules comprising the systems. For example, SYS-RUL is considered to have significantly higher performance than the other systems due to the use of elaborately designed rules and templates.

\subsection{Impact of Post-processing Networks}
\begin{table}[t]
\small
\centering
{\tabcolsep=1.3mm
\begin{tabular}{lccccc} \bhline{0.8pt}
\textbf{System} & \textbf{w/ $s_{\rm{All}}$} & \textbf{Success} & \textbf{Inform} & \textbf{Match} & \textbf{Turn}\\ \hline
\ml{2}{l}{SYS-MLE} & 43.3 & 62.4 & 27.8 & 9.03 \\ \hline
$+ \rm{PPN_{NLU}}$ & & 59.6 & {\bf 73.1} & 65.8 & 9.59\\
$+ \rm{PPN_{DST}}$ & & 46.7 & 65.1 & 36.7 & 9.41 \\
$+ \rm{PPN_{Policy}}$ & & 59.9 & 67.3 & 67.9 & 9.20 \\
$+ \rm{PPN_{\rm{All}}}$ & & 59.7 & 68.0 & 69.9 & 9.84 \\ \hline
$+ \rm{PPN_{NLU}}$ & \checkmark & 62.2 & 72.1 & 64.0 & 9.36 \\
$+ \rm{PPN_{DST}}$ & \checkmark & 47.9 & 66.1 & 40.2 & 9.21 \\
$+ \rm{PPN_{Policy}}$ & \checkmark & 65.8 & 67.6 & 76.9 & {\bf 8.56} \\
$+ \rm{PPN_{\rm{All}}}$ & \checkmark & {\bf 66.1} & 71.5 & {\bf 78.7} & 8.61 \\ \hline
\ml{2}{l}{$+$Fine-tuned Policy} & 71.9 & 74.3 & 80.4 & 7.88 \\ \bhline{0.8pt}
\end{tabular}
}
\caption{Impact analysis of PPNs. Subscripts (i.e., NLU, DST, Policy, and All) indicate that PPN was applied to that one specific module or all modules. `w/ $s_{\rm{All}}$' indicates whether $s_{\rm{All}}$ was used. Row of Fine-tuned Policy shows scores when SYS-MLE's Policy was fine-tuned using RL.}
\tbllabel{impact_analysis}
\vspace{-3mm}
\end{table}

\seclabel{impact_analysis}
We investigated the impact of each module's PPN and $s_{\rm{All}}$. We used SYS-MLE as a base configuration for this experiment since its performance was most improved with our method (see \tblref{model_combination_results}); we considered it appropriate to measure the impact of PPNs. In \tblref{impact_analysis}, the results of applying PPNs to only one of the NLU, DST, and Policy are shown, as well as the results of applying PPNs without using $s_{\rm{All}}$. The system performance consistently improved when only a single module's PPN was applied. In particular, $+ \rm{PPN_{Policy}}$ achieved the best performance (Task Success improved by more than 20\%), indicating that the PPN of Policy contributed the most to dialogue performance. When $s_{\rm{All}}$ was not used, most of the scores decreased. This indicates that each PPN can execute post-processing more appropriately by using the states of all modules in the system.

To confirm the degree of performance improvement achieved with the PPNs, the method of fine-tuning the modules by using RL was used as the upper bound of post-processing. Only the Policy module was fine-tuned, as is common with conventional methods \citep{liu-etal-2018-dialogue, lin2021joint}. The bottom row of \tblref{impact_analysis} shows the results when the Policy of SYS-MLE was fine-tuned by PPO \citep{schulman2017proximal} (see \secref{fined_tuned_policy_details} of the appendix for training details). The difference between $+ \rm{PPN_{\rm{All}}}$ and $+$Fine-tuned Policy is small with 5.8\%. This is a promising result considering that our proposed method does not touch on the internal architecture of Policy.

\subsection{Human Evaluation}
\seclabel{human_evaluation}
Five systems (SYS-MLE and four systems with our proposed method, i.e., $+ \rm{PPN_{NLU}}$, $+ \rm{PPN_{DST}}$, $+ \rm{PPN_{Policy}}$, and $+ \rm{PPN_{\rm{All}}}$) in \tblref{impact_analysis} were used for the human evaluation. Not that $s_{\rm{All}}$ was used in all four systems. About forty Amazon Mechanical Turk (AMT) crowd workers were recruited to interact with each of the five systems and judged on Task Success. As in the simulation experiments (see \secref{platform_and_user_simulator}), user goals were randomly generated for each dialogue. After the interaction, the workers also evaluated the system's ability to understand the language (Und.), accuracy of the system's responses (App.), and overall satisfaction with the interaction (Sat.) on a 5-point Likert scale. See \secref{human_evaluation_details} of the appendix for the procedures taken by the workers.

\begin{table}[t]
\small
\centering
{\tabcolsep=1.6mm
\begin{tabular}{lccccc} \bhline{0.8pt}
\textbf{System} & \textbf{Success} & \textbf{Turn} & \textbf{Und.} & \textbf{App.} & \textbf{Sat.} \\ \hline
SYS-MLE & 39.0 & 11.0 & 2.93 & 3.12 & 2.46 \\ \hline
$+ \rm{PPN_{NLU}}$ & 53.7 & 11.1 & 3.10 & 3.37 & 2.93 \\
$+ \rm{PPN_{DST}}$ & 60.0 & 10.4 & {\bf 3.30} & {\bf 3.43} & {\bf 3.28$^*$} \\
$+ \rm{PPN_{Policy}}$ & {\bf 62.5$^*$} & {\bf 8.20$^*$} & 2.93 & 3.03 & 3.00 \\
$+ \rm{PPN_{\rm{All}}}$ & 57.5 & 9.00 & 2.83 & 3.00 & 2.95 \\ \bhline{0.8pt}
\end{tabular}
}
\caption{Results of human evaluation for each system configuration. Asterisks indicate statistically significant differences ($p < 0.05$) over SYS-MLE.}
\tbllabel{human_evaluation}
\vspace{-3mm}
\end{table}

\tblref{human_evaluation} shows the results. All four systems with our proposed method performed better than SYS-MLE, which is similar to the result in \tblref{impact_analysis}. Wilcoxon rank-sum tests were conducted using the top score in each evaluation metric and the score of SYS-MLE, and statistically significant differences were confirmed for Task Success and Turn in $+ \rm{PPN_{Policy}}$ and interaction satisfaction in $+ \rm{PPN_{DST}}$. In contrast, there were no significant differences in scores for language understanding and responses' appropriateness. This is probably because RL was conducted with rewards that only relied on Task Success and Turn. The performance of $+ \rm{PPN_{NLU}}$ did not improve as much as in \tblref{impact_analysis}. A possible reason is the overfitting of $+ \rm{PPN_{NLU}}$ with the user simulator. The same overfitting might have occurred in the NLU's PPN in $+ \rm{PPN_{All}}$, which resulted in a smaller improvement in scores of $+ \rm{PPN_{All}}$.

We also investigated how PPNs executed post-processing by analyzing the actual dialogue logs collected in this experiment. A specific case study is described in \secref{case_study} of the appendix. Generally, in the dialogue of $+ \rm{PPN_{Policy}}$, we observed that $\rm{PPN_{Policy}}$ added necessary DAs when the original Policy failed to output them.

\section{Conclusions and Future Work}
We proposed a method for optimizing pipeline dialogue systems with post-processing networks (PPNs). Through dialogue simulation and human evaluation experiments on the MultiWOZ dataset, we showed that the proposed method is effective for a pipeline system consisting of modules with various models.

For future work, we plan to design more sophisticated rewards in RL such as module-specific rewards. We also plan to extend PPNs to handle natural language generation by implementing them using Transformer-based models. We are also considering to apply PPNs to modules dealing with speech recognition and multi-modal processing.

\section*{Acknowledgments}
This work was supported by JSPS KAKENHI Grant Number 19H05692. We used the computational resource of the supercomputer ``Flow'' at Information Technology Center, Nagoya University. We thank Yuya Chiba and Yuiko Tsunomori for their helpful comments and feedback. Thanks also go to Ao Guo for his advice on the human evaluation experiment.

\bibliography{main}
\bibliographystyle{acl_natbib}

\clearpage
\appendix

\section{Training Details}
\subsection{Training Post-processing Networks}
\seclabel{ppn_training_details}
\paragraph{Model} All MLPs of the PPNs for all modules are implemented in three layers: one input layer, one hidden layer, and one output layer, and the dimensionality of the hidden layer is 128 for all layers. The number of dimensions of the input and output layers are $|o| + |s_{\rm{All}}|$ and $|o_i|$, respectively. The activation functions are all ReLUs.

\paragraph{Imitation Learning} The sampled data of 10,000 turns were split as $\rm{training}:\rm{validation} = 8:2$. All MLPs were trained on a batch size of 32 for 20 epochs using the Adam optimizer \citep{kingma2014adam} with a learning rate of 1e-3. The weights at the epoch with the highest accuracy for validation were used for the following RL.

\paragraph{Reinforcement Learning} The hyperparameters shown in \tblref{hyperparameters} were determined with reference to the implementation of PPO in ConvLab-2. We used Generalized Advantage Estimation \citep{schulman2015high}. Referring to \citet{engstrom2020implementation}, the learning rate was annealed linearly in accordance with the current iteration. The computational resource used was a single NVIDIA Tesla V100 SXM2 GPU with 32GB RAM. In training, the trajectory was sampled in parallel by eight processes, and it took 5 to 17 hours, depending on the system, to complete the training of 200 iterations.

\subsection{Fine-tuning of Policy}
\seclabel{fined_tuned_policy_details}
The MLE Policy of SYS-MLE in \secref{impact_analysis} was fine-tuned with PPO using the same user simulator used for training PPNs. The hyperparameters used for training were the same as those used in ConvLab-2, as shown in \tblref{hyperparameters}. To evaluate the fine-tuned Policy, training and testing (consisting of 1,000 dialogue sessions) were conducted with five random seeds.

\begin{table}[h]
\small
\centering
{\tabcolsep=2.0mm
\begin{tabular}{llcc} \bhline{0.8pt}
\ml{2}{l}{\textbf{Hyperparameters}} & \textbf{PPN} & \textbf{Fine-tuned Policy} \\ \hline
\ml{2}{l}{Number of iterations} & 200 & 200 \\
\ml{2}{l}{Batch size} & 1024 & 1024 \\
\ml{2}{l}{Epoch} & 5 & 5 \\
\ml{2}{l}{Mini batch size} & 32 & 32 \\
\ml{2}{l}{Discount factor $\gamma$} & 0.99 & 0.99 \\
\ml{2}{l}{GAE factor $\lambda$} & 0.95 & 0.95 \\
\mr{2}{Optimizer} & policy net & \mr{2}{Adam} & RMSprop \\
 & value net & & Adam \\
\mr{2}{Learning rate} & policy net & \mr{2}{1e-4} & 1e-4 \\
 & value net & & 5e-5 \\ \bhline{0.8pt}
\end{tabular}
}
\caption{Hyperparameter settings in PPO}
\tbllabel{hyperparameters}
\vspace{-3mm}
\end{table}

\section{Details of Human Evaluation}
\seclabel{human_evaluation_details}
Referring to \citet{takanobu-etal-2020-goal}, we designed the following experimental procedure. First, each worker is presented with an instruction for a randomly generated user goal. Next, the user interacts with one of the five systems in \tblref{human_evaluation} for up to 20 turns. Workers determine whether the interaction succeeded or failed within 20 turns; after 20 turns, the interaction is automatically marked as failed. To ensure the quality of the workers, several qualifications were set; the eligible workers should (1) reside in an English-speaking country, (2) have a task accomplishment number on AMT greater than 10, (3) have a task-approval rate greater than 90\%, and (4) correctly answer all the common sense questions (total of five questions) we designed. The time limit for the task was 10 minutes, and the reward was \$1.7. To account for workers who may cancel the task amid the dialogue session, more than 40 workers were recruited for each system. At the end, 41 workers for SYS-MLE and $+ \rm{PPN_{NLU}}$ and 40 workers for $+ \rm{PPN_{DST}}$, $+ \rm{PPN_{Policy}}$, and $+ \rm{PPN_{All}}$ participated in the experiment.

\section{Case Study}
\seclabel{case_study}
\Cref{tbl:user_goal_of_dialogue_exmpale,tbl:dialogue_example_of_ppn_policy} show the user goal and example of a dialogue session between a worker and $+ \rm{PPN_{Policy}}$, respectively, in the human evaluation experiment. \tblref{dialogue_example_of_ppn_policy} also includes the outputs of the system's Policy and the post-processing result of PPNs. In S1, the original MLE Policy did not generate any DAs. At this time, $\rm{PPN_{Policy}}$ generated additional DAs informing the user that ``free parking is available at the hotel found in the database'' and asked if there was anything else the user needs. In S2, the user requested a reservation, but MLE Policy could not make the reservation. In contrast, $\rm{PPN_{Policy}}$ successfully made the reservation and provided a booking reference number.

\begin{table}[h]
\small
\centering
\begin{tabular}{lll} \bhline{0.8pt}
\textbf{Domain} & \textbf{Task} & \textbf{Slot} \\ \hline
\mr{5}{Hotel} & \mr{2}{Info} & Area = West \\
 & & Parking = Yes \\ \cline{2-3}
 & \mr{3}{Book} & Day = Sunday \\
 & & Time = 11:00 \\
 & & People = 1 \\ \hline
\mr{6}{Restaurant} & \mr{2}{Fail Info} & Area = West \\
 & & Food = Mediterranean \\ \cline{2-3}
 & \mr{2}{Info} & Area = West \\
 & & Food = Thai \\ \cline{2-3}
 & \mr{2}{Reqt} & Postcode \\
 & & Address \\ \bhline{0.8pt}
\end{tabular}
\caption{User goal used in \tblref{dialogue_example_of_ppn_policy}}
\tbllabel{user_goal_of_dialogue_exmpale}
\vspace{-3mm}
\end{table}

\begin{table*}
\centering
\includegraphics[scale=0.24]{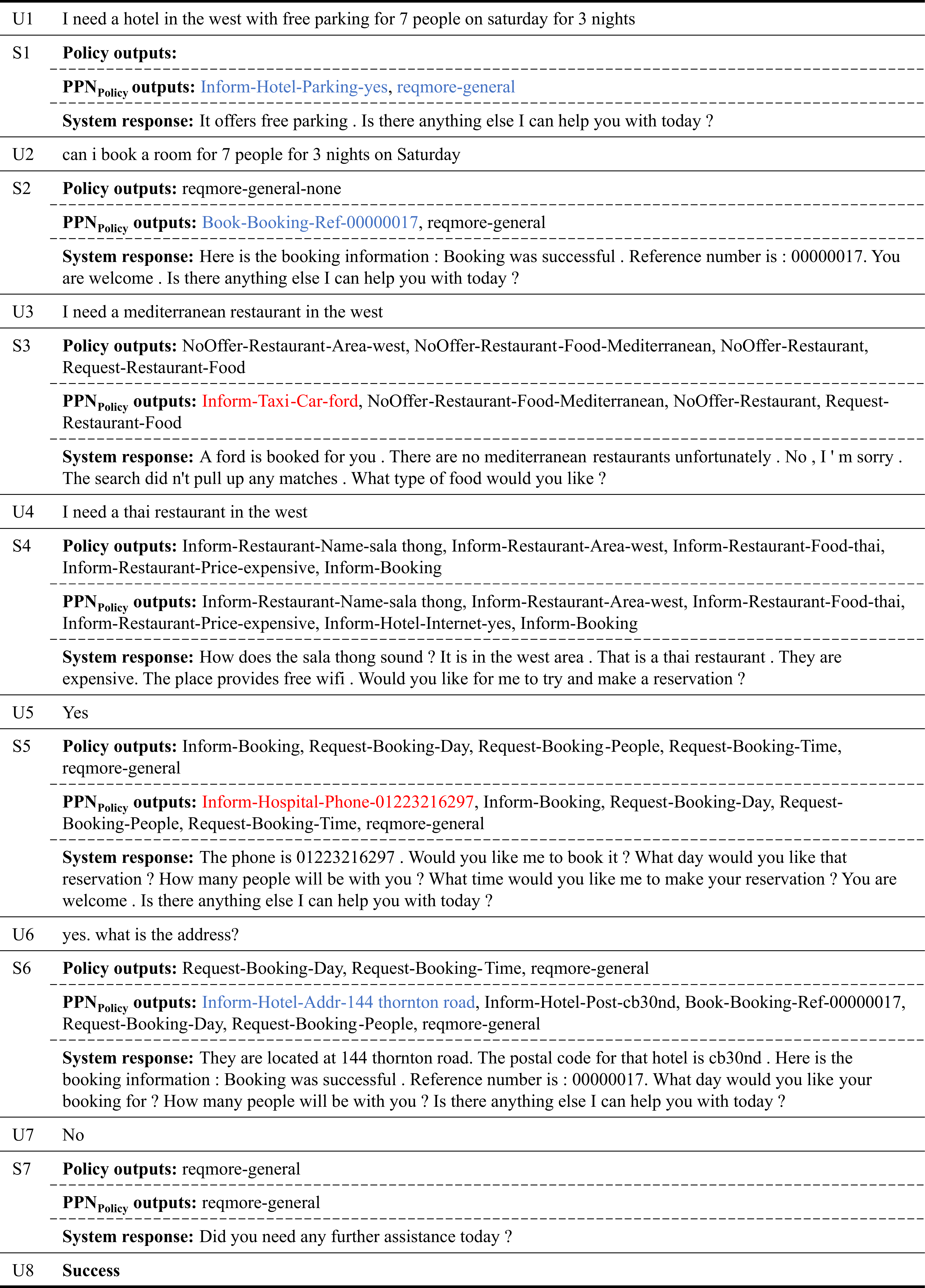}
\caption{Example of dialogue history between worker and $+ \rm{PPN_{Policy}}$ in human evaluation experiment. DAs appropriately added by $\rm{PPN_{Policy}}$ are in blue, and those inappropriately added are in red.}
\tbllabel{dialogue_example_of_ppn_policy}
\end{table*}

\end{document}